\documentclass[11pt]{article}

\usepackage{acl}

\usepackage{array}
\usepackage{times}
\usepackage{latexsym}
\usepackage{tabularx} 
\usepackage[cmyk]{xcolor}%
\definecolor{my_blue}{cmyk}{0.04, 0.02, 0, 0}
\definecolor{cvpr_blue}{cmyk}{0.72,0.34,0,0.26}
\definecolor{my_green}{cmyk}{0.04, 0, 0.06, 0.02}
\definecolor{acl_green}{cmyk}{0.74,0,0.21,0.40}
\definecolor{mycyan}{cmyk}{0.065, 0, 0, 0}
\definecolor{thu_purple}{cmyk}{0.65,1,0,0.2}

\usepackage{pifont} %
\usepackage{colortbl}
\usepackage[most]{tcolorbox}

\usepackage{algorithm}
\usepackage{algorithmic}
\usepackage{multirow}
\usepackage{booktabs}
\usepackage{newfloat}
\usepackage{listings}
\usepackage[T1]{fontenc}

\usepackage[utf8]{inputenc}

\usepackage{microtype}

\usepackage{graphicx}

\definecolor{my_blue}{cmyk}{0.04, 0.02, 0, 0}
\definecolor{cvpr_blue}{cmyk}{0.72,0.34,0,0.26}
\definecolor{my_yellow}{cmyk}{0.04, 0, 0.06, 0.02}
\definecolor{acl_green}{cmyk}{0.74,0,0.21,0.40}
\definecolor{mycyan}{cmyk}{0.065, 0, 0, 0}

\definecolor{my_deep_yellow}{cmyk}{0, 0.008, 0.1, 0.08}
\definecolor{my_yellow}{cmyk}{0, 0.002, 0.04, 0.020}

\usepackage{pifont} %
\usepackage{colortbl}
\usepackage[most]{tcolorbox}

\usepackage{algorithm}
\usepackage{algorithmic}
\usepackage{multirow}
\usepackage{booktabs}
\definecolor{cvprblue}{rgb}{0.21,0.49,0.74}

\usepackage{inconsolata}
\usepackage{xcolor} 

\hypersetup{
    colorlinks=true,      
    linkcolor=acl_green,       
    citecolor=acl_green,    
    urlcolor=acl_green,      
}
\title{MCGA: A Multi-task Classical Chinese Literary Genre Audio Corpus}

\begin{document}

\author{Yexing Du$^{1,2}$\footnotemark[1] \quad Kaiyuan Liu$^{1,2}$\footnotemark[1] \quad Bihe Zhang$^{3}$ \quad Youcheng Pan$^{2}$ \quad Bo Yang$^{2}$ \quad Liangyu Huo$^{4}$ \\ \textbf{Xiyuan Zhang$^{4}$ 
 \quad Jian Xie$^{4}$ \quad Daojing He$^{1}$\footnotemark[2] \quad Yang Xiang$^{2}$\footnotemark[2] \quad Ming Liu$^{1,2}$\footnotemark[2] \quad Bing Qin$^{1,2}$} \\ 
$^1$Harbin Institute of Technology \quad $^2$Pengcheng Laboratory\\ \quad $^3$South China University of Technology \quad $^4$Du xiaoman \\ 
\texttt{{yxdu@ir.hit.edu.cn,}  {1171000408@stu.hit.edu.cn}}\\
\texttt{{hedaojinghit@163.com,} {xiangy@pcl.ac.cn,} {mliu@ir.hit.edu.cn}}}

\maketitle
\renewcommand{\thefootnote}{\fnsymbol{footnote}} 
\footnotetext[1]{Equal contribution.} 
\footnotetext[2]{Corresponding authors.}
\renewcommand{\thefootnote}{\arabic{footnote}}
\begin{abstract}
With the rapid advancement of Multimodal Large Language Models (MLLMs), their potential has gained significant attention in Chinese Classical Studies (CCS). While existing research primarily focuses on text and visual modalities, the audio corpus within this domain remains largely underexplored. To bridge this gap, we introduce the \textbf{Multi-task Classical Chinese Literary Genre Audio Corpus (MCGA)}, a 119-hour corpus comprising 22,000 audio samples. It encompasses a diverse range of literary genres across six tasks: Automatic Speech Recognition (ASR), Speech-to-Text Translation (S2TT), Speech Emotion Captioning (SEC), Spoken Question Answering (SQA), Speech Understanding (SU), and Speech Reasoning (SR). 
Through the evaluation of ten MLLMs, our experimental results demonstrate that current MLLMs still face substantial challenges on the MCGA test set. Furthermore, we introduce a domain-specific metric for SEC and a metric to measure the consistency between speech and text capabilities. We release MCGA to the public to facilitate the development of more robust MLLMs.\footnote{MCGA Corpus: \url{https://github.com/yxduir/MCGA}}
\end{abstract}

\section{Introduction}
The development of Multimodal Large Language Models (MLLMs)~\cite{li_unimoe,li2025perception,li2025uni} has significantly advanced Chinese Classical Studies (CCS). These models support multimodal inputs, providing powerful capabilities for interpreting ancient texts, which in turn enhances cultural preservation~\cite{zhang-etal-2025-mllms}. However, while most existing research focuses on textual~\cite{cao2024wenmind} or visual~\cite{liu2025mcs} modalities, the auditory dimension of CCS remains largely unexplored due to a lack of high-quality, domain-specific audio corpora.
\begin{figure}[t]
\includegraphics[width=\columnwidth]{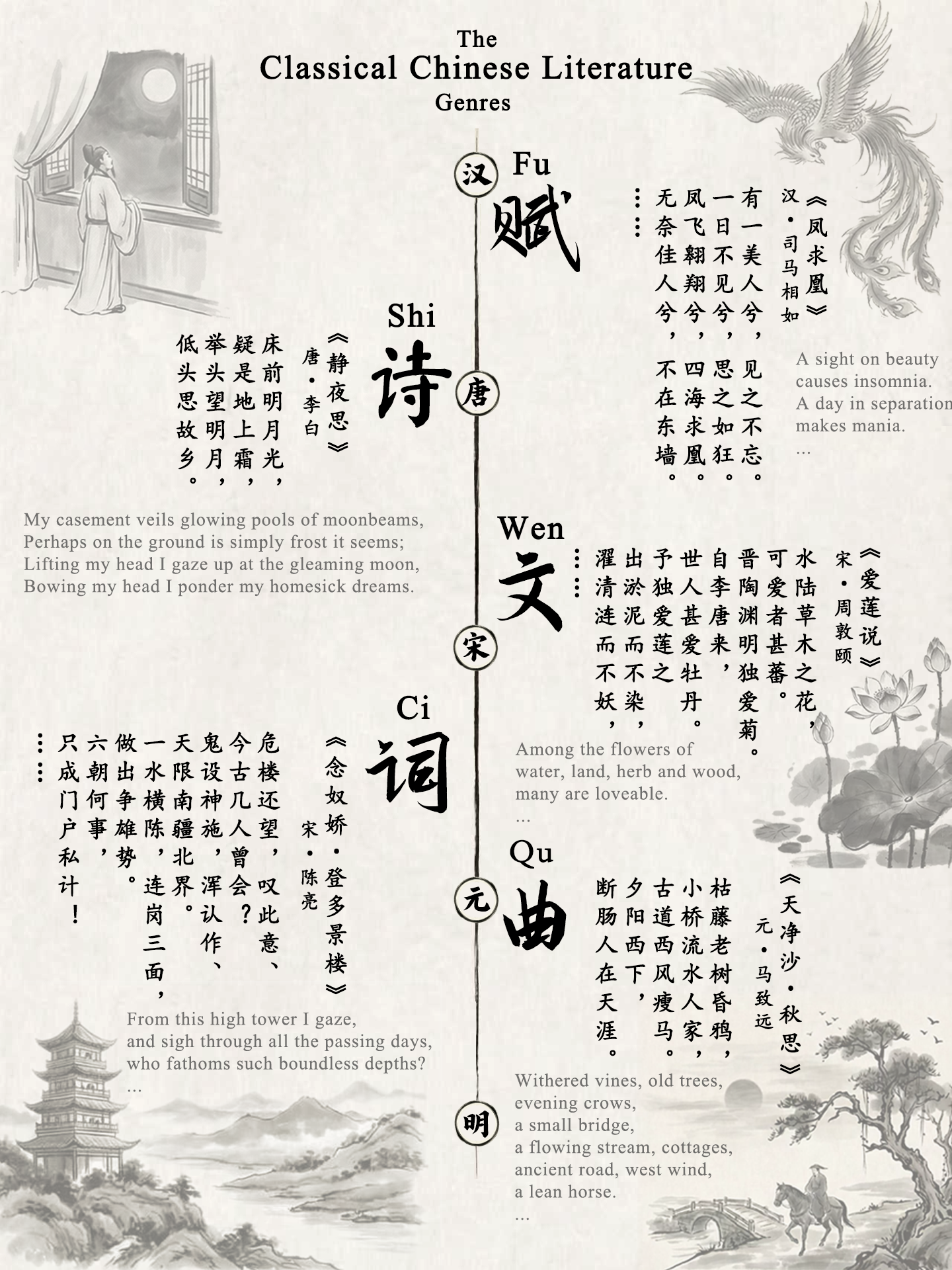}
\caption{\textbf{Timeline of the Golden Age for Classical Chinese Literary Genres}: \textit{Fu} (Rhapsody), \textit{Shi} (Poetry), \textit{Wen} (Prose), \textit{Ci} (Lyric), and \textit{Qu} (Song).}
\label{intro}
\end{figure}
\begin{figure*}[t] \centering
    \includegraphics[width=\textwidth]{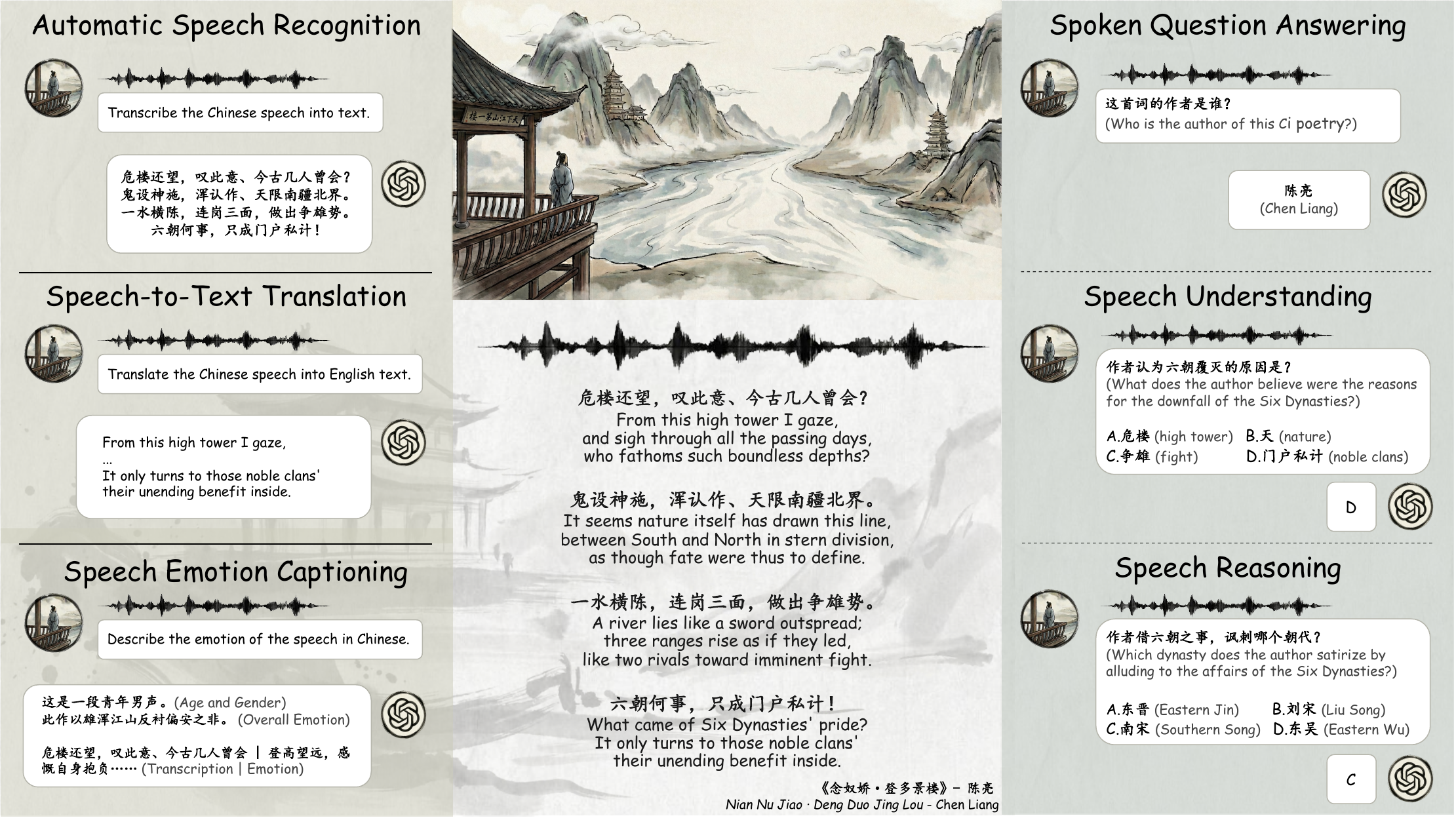}
\caption{\textbf{Examples from the MCGA Corpus.} The corpus covers six core speech tasks (ASR, S2TT, SEC, SQA, SU, SR). Leveraging its parallel speech-text data, it also supports four text tasks: Machine Translation (MT), Question Answering (QA), Language Understanding (LU), and Language Reasoning (LR).}
\label{fig:sample}
\end{figure*}

To bridge this critical gap, we introduce the \textbf{Multi-task Classical Chinese Literary Genre Audio Corpus (MCGA)}, a comprehensive resource designed to catalyze audio-centric research in CCS. As illustrated in Figure~\ref{intro}, MCGA encompasses five primary literary genres: \textit{Fu}, \textit{Shi}, \textit{Wen}, \textit{Ci}, and \textit{Qu}. The corpus consists of 22,000 audio samples, totaling 119 hours of recorded content. To ensure cultural and linguistic authenticity, the data were recorded by native speakers in standard Mandarin Chinese. Crucially, all audio samples include explicit copyright transfers, thereby resolving long-standing Intellectual Property Rights (IPR) challenges in open-source CCS audio datasets.

The MCGA corpus offers two primary advantages:
(1) \textbf{Task Diversity}: As illustrated in Figure~\ref{fig:sample}, the corpus supports 6 diverse speech-centric tasks, including Automatic Speech Recognition (ASR), Speech-to-Text Translation (S2TT), Speech Emotion Captioning (SEC), Spoken Question Answering (SQA), Speech Understanding (SU), and Speech Reasoning (SR), alongside four integrated text tasks. 
(2) \textbf{Literary Genre Diversity}: It encompasses 5 major literary genres spanning 11 historical periods, forming a total of 37 distinct period-genre categories and covering a comprehensive collection of 4,497 literary works.

We evaluated 10 representative MLLMs, including 2 closed-source and 8 open-source models. 
Experimental results indicate that current MLLMs still have significant room for improvement in the CCS field. Notably, even the top-performing model, Qwen3-Omni~\cite{xu2025qwen3}, scored below 60 on complex tasks such as SEC. 
Besides, we introduce the Emotion Caption Fidelity (ECF) metric tailored for literary SEC task, along with the Cross-Modal Consistency (CMC) metric to quantify the alignment between a model's auditory and textual reasoning. Furthermore, the substantial performance gains achieved through training underscore the superior quality of the MCGA corpus.

\begin{table*}[t]
\centering
\small
    \renewcommand{\arraystretch}{1.2} 

\resizebox{\textwidth}{!}{

\setlength{\tabcolsep}{6pt} 
\begin{tabular}{lccccccccccc}
\toprule

\multirow{2}{*}{\textbf{Dataset} (Text\textcolor{acl_green}{\ding{61}} / Image\textcolor{blue}{\ding{67}} / Audio\textcolor{thu_purple}{\ding{71}})}   & \multirow{2}{*}{\textbf{Modality}}& \multirow{2}{*}{\textbf{Domain}}& \multirow{2}{*}{\textbf{Scale}} & \multirow{2}{*}{\textbf{License}} & \multirow{2}{*}{\textbf{Copyright}}\\ 
 &&&  &  & \\
\midrule

\rowcolor{my_green}ACLUE~\cite{zhang-li-2023-large} &  \textcolor{acl_green}{\ding{61}} & CCS & 4,967&CC BY-NC-4.0  &\\
\rowcolor{my_green}CCLUE~\cite{wang-etal-2023-rethinking} &  \textcolor{acl_green}{\ding{61}} & CCS & 36,319  &Apache-2.0 &\\
\rowcolor{my_green}WYWEB~\cite{zhou-etal-2023-wyweb} &  \textcolor{acl_green}{\ding{61}} & CCS & 69,700 &- &\\
\rowcolor{my_green}WenMind~\cite{cao2024wenmind}  & \textcolor{acl_green}{\ding{61}} & CCS & 4,875  & CC BY-NC-SA-4.0  & \\
\rowcolor{my_green}TianWen~\cite{pei2025tianwen}  & \textcolor{acl_green}{\ding{61}} & CCS &4,000  & MIT  & \\ \midrule
\rowcolor{my_yellow}CII-Bench~\cite{zhang-etal-2025-mllms} & \textcolor{acl_green}{\ding{61}} /   \textcolor{blue}{\ding{67}} & General& 698  & Apache-2.0  &  \\
\rowcolor{my_yellow}FoodieQA~\cite{foodieqa2024}  & \textcolor{acl_green}{\ding{61}} /   \textcolor{blue}{\ding{67}} & Food & 389& CC BY-NC-ND-4.0 &  \textcolor{blue}{\ding{67}}\\
\rowcolor{my_yellow}Oracle-Bench~\cite{qiao-etal-2025-v}  & \textcolor{acl_green}{\ding{61}} /   \textcolor{blue}{\ding{67}} & CCS& 2,834 & - &  \\
\rowcolor{my_yellow}Paint4Poem~\cite{li2021paint4poem}  & \textcolor{acl_green}{\ding{61}} /   \textcolor{blue}{\ding{67}}&   CCS &93,153   &  Github  &  \\
\rowcolor{my_yellow}MCS-Bench~\cite{liu2025mcs}  & \textcolor{acl_green}{\ding{61}} /   \textcolor{blue}{\ding{67}} & CCS& 6,500 & CC BY-NC-SA-4.0 &  \\
\midrule
\rowcolor{my_yellow}\multirow{-1}{*}{\textbf{MCGA} (ours)}&\multirow{-1}{*}{\textcolor{acl_green}{\ding{61}} /   \textcolor{thu_purple}{\ding{71}}}&\multirow{-1}{*}{CCS}&\multirow{-1}{*}{22,000}&\multirow{-1}{*}{CC BY-NC-SA-4.0}&\textcolor{thu_purple}{\ding{71}}\\
\bottomrule
\end{tabular}
}
\caption{\textbf{Comparison of MCGA with Existing Chinese Cultural Datasets}. MCGA is the first large-scale, fully copyrighted classical Chinese literary audio corpus for MLLMs (119 hours). All recordings are sourced directly from original creators with full copyright transfer, highlighting our commitment to Intellectual Property Rights (IPR) protection in Chinese Classical Studies (CCS) research. Copyright here refers to the original authorship and ownership of the raw audio and image materials.}
\label{tab:compare_prior}
\end{table*}

\newpage

Our primary contributions are as follows:
\begin{itemize}
    \setlength{\itemsep}{2pt}
    \item \textbf{MCGA Corpus:} We present MCGA, the first large-scale (119 hours), open-source, and fully copyrighted audio corpus dedicated to classical Chinese literature studies. This resource effectively bridges the gap in high-quality audio datasets for this domain.
    
    \item \textbf{Evaluation Framework:} We establish a comprehensive evaluation framework centered on MCGA, comprising 6 multifaceted tasks: ASR, S2TT, SEC, SQA, SU, and SR. This enables a rigorous investigation into the capabilities of MLLMs.
    
  \item \textbf{Evaluation Metrics:} We introduce two novel evaluation metrics: the ECF metric tailored for literary SEC task, and a CMC metric designed to assess the alignment and quantify the gap between auditory and textual modalities.

    \item \textbf{Empirical Analysis:} We evaluate 10 MLLMs to identify performance bottlenecks in the classical Chinese literature domain. Besides, we demonstrate MCGA's high utility as a training resource, where fine-tuning yields substantial performance breakthroughs.
\end{itemize}

\newpage

\section{Related Works}
\subsection{Chinese Cultural Datasets}
The landscape of Chinese cultural evaluation spans many domains~\cite{DOI:10.3724/2096-7004.di.2025.0084}. ACLUE~\cite{zhang-li-2023-large} and WYWEB~\cite{zhou-etal-2023-wyweb} establish large-scale benchmarks for Classical Chinese and ancient literature, focusing on linguistic understanding. Complementarily, CCLUE~\cite{wang-etal-2023-rethinking} rethinks cultural evaluation across broader contexts. In the multimodal sphere, FoodieQA~\cite{foodieqa2024} and CII-Bench~\cite{zhang-etal-2025-mllms}  probe culinary arts and figurative reasoning, respectively, highlighting a shift toward assessing complex cultural heritage and everyday traditions.

\subsection{Chinese Classical Studies Datasets}
Recent benchmarks deepen the evaluation of Chinese classical heritage through diverse methodologies~\cite{WOS:001551642700001}. WenMind~\cite{cao2024wenmind} assesses deep cultural cognition and mentalities, while TianWen~\cite{pei2025tianwen} provides specialized assessment for traditional scriptures and historical knowledge. Advancing into multimodality, Oracle-Bench~\cite{qiao-etal-2025-v} evaluates ancient script deciphering, whereas Paint4Poem~\cite{li2021paint4poem} bridges classical poetry with visual synthesis. MCS-Bench~\cite{liu2025mcs} offers a framework for multimodal classical studies. However, few of these benchmarks or datasets contain the parallel speech of the classical Chinese literature.

\newpage

\section{MCGA Corpus}
\subsection{Overview}
We introduce MCGA, a comprehensive corpus designed to promote audio-centric research in CCS. This section briefly outlines the construction, the human recording process, the subsequent quality control and the statistics of MCGA.

\subsection{Data Construction}
\paragraph{Data Collection and Preprocessing.}
Classical Chinese literature and corresponding Pinyin were sourced from the web. All works are in the public domain (created over 150 years ago). Following cleaning, texts were segmented to restrict recording duration to 30 seconds.

\paragraph{Text Data Construction.}
Subsequently, we leverage DeepSeek-V3.2~\cite{guo2025deepseek} to generate question-answer pairs for the text of each clip, with access to the full literary context. This process covers a variety of speech-related tasks, including S2TT, SEC, SQA, SU, and SR.

\paragraph{Text Data Check.}
The generated question–answer pairs are subjected to trio validation using DeepSeek-V3.2, GPT-5-mini~\cite{gpt5}, and Gemini-3-Flash~\cite{gemini3flash}, through which pairs that fail to pass the verification are filtered out. The test and validation sets underwent human verification to ensure data quality.

\newpage

\begin{figure}[h]
\centering
\includegraphics[width=\linewidth]{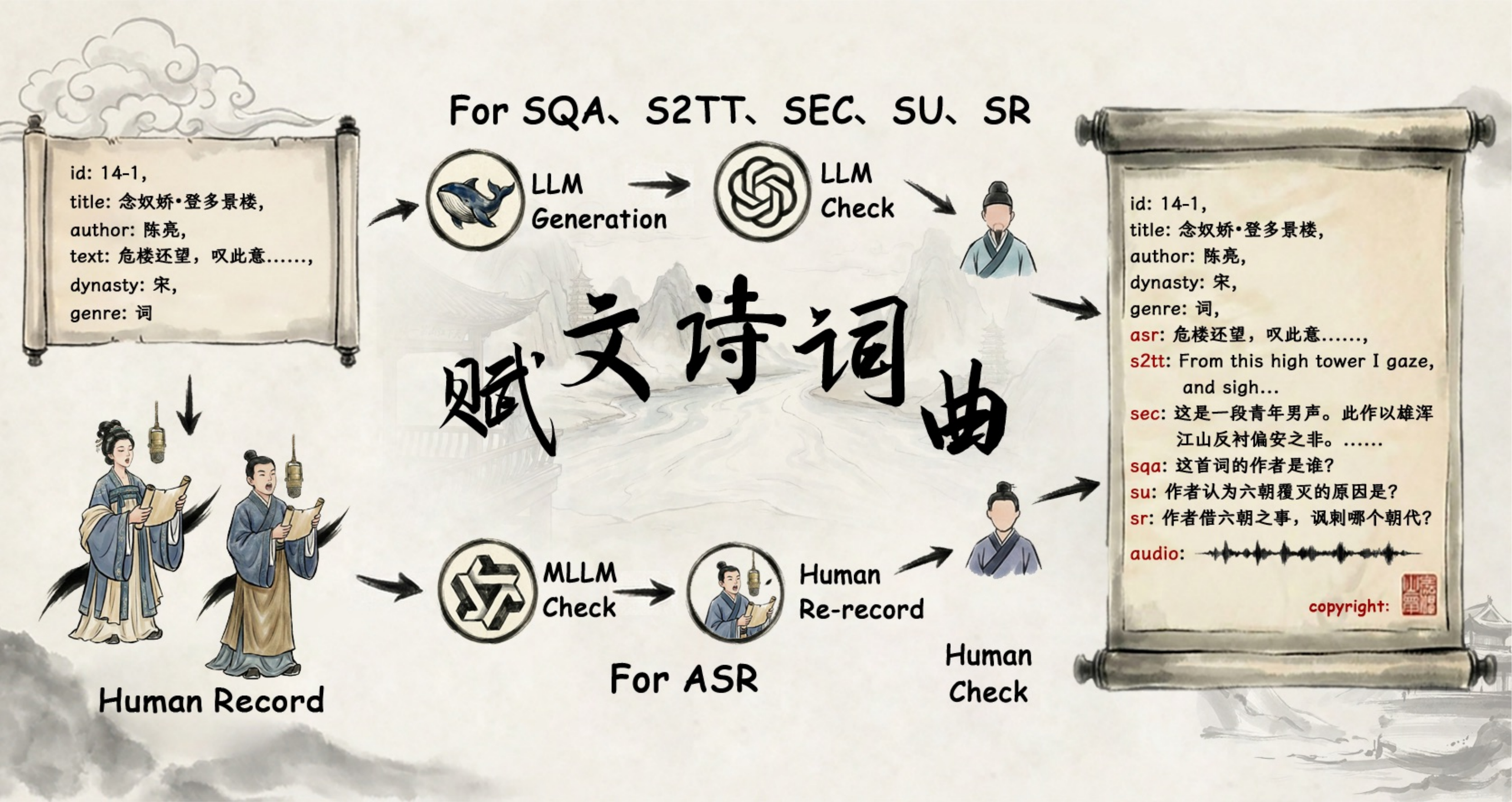}
\caption{\textbf{MCGA Corpus Construction.} Initially comprising only metadata such as titles, authors, and texts, the MCGA corpus is expanded through human recording, LLM generation, and rigorous verification. Then, it supports six speech tasks: ASR, S2TT, SEC, SQA, SU, and SR. We provide a detailed example of the SEC task in Figure~\ref{case_sec}.}
\label{fig:procedure}
\end{figure}

\subsection{Human Recording}
\paragraph{Volunteer Demographics.}
We recruited 28 native speakers (13 males and 15 females, aged 18–40) to record the texts via a dedicated private website. All participants have good educational backgrounds, half of whom are Chinese majors.

\paragraph{Recording Protocol.}
We explicitly stated the recording guidelines to the volunteers to ensure high-quality audio acquisition and emotional consistency across the dataset, as follows:


\begin{tcolorbox}[
    colback=my_yellow,
    colframe=my_deep_yellow,
    title=\textbf{{\color{black}Volunteer Guidelines}},
    center title,
    boxrule=1pt,
    arc=2mm,
    left=-2pt,
    right=4pt,
    top=4pt,
    bottom=4pt,
]
\fontsize{9pt}{10.5pt}\selectfont
\begin{itemize}
    \setlength{\itemsep}{2pt}
    \setlength{\parskip}{0pt}
    \setlength{\parsep}{0pt}
    \item use a tone that matches the emotion of the text.

    \item can access the pinyin for all Chinese characters.

    \item ensure the recording environment is quiet.

    \item each clip must be read in 30 seconds.

    \item each clip is read by at least 1 male and 1 female.

    \item clips from the same work are sent to the same person.
\end{itemize}
\end{tcolorbox}

\subsection{Audio Quality Check}
\paragraph{MLLM Check.}
We employed a dual-stage speech recognition verification process using Qwen and Whisper~\cite{whisper} models to identify samples with significant errors, which were subsequently re-recorded by the volunteers.

\paragraph{Human Check.}
We recruited 6 data quality inspection volunteers to verify the validation and test sets. The inspectors were instructed to score the samples. Low quality samples (pronunciation error or presence of background noise) were removed from the sets.

Both recording volunteers and quality inspection volunteers signed labor agreements and were compensated with reasonable remuneration.

\begin{figure}[t]
\centering
\includegraphics[width=\linewidth]{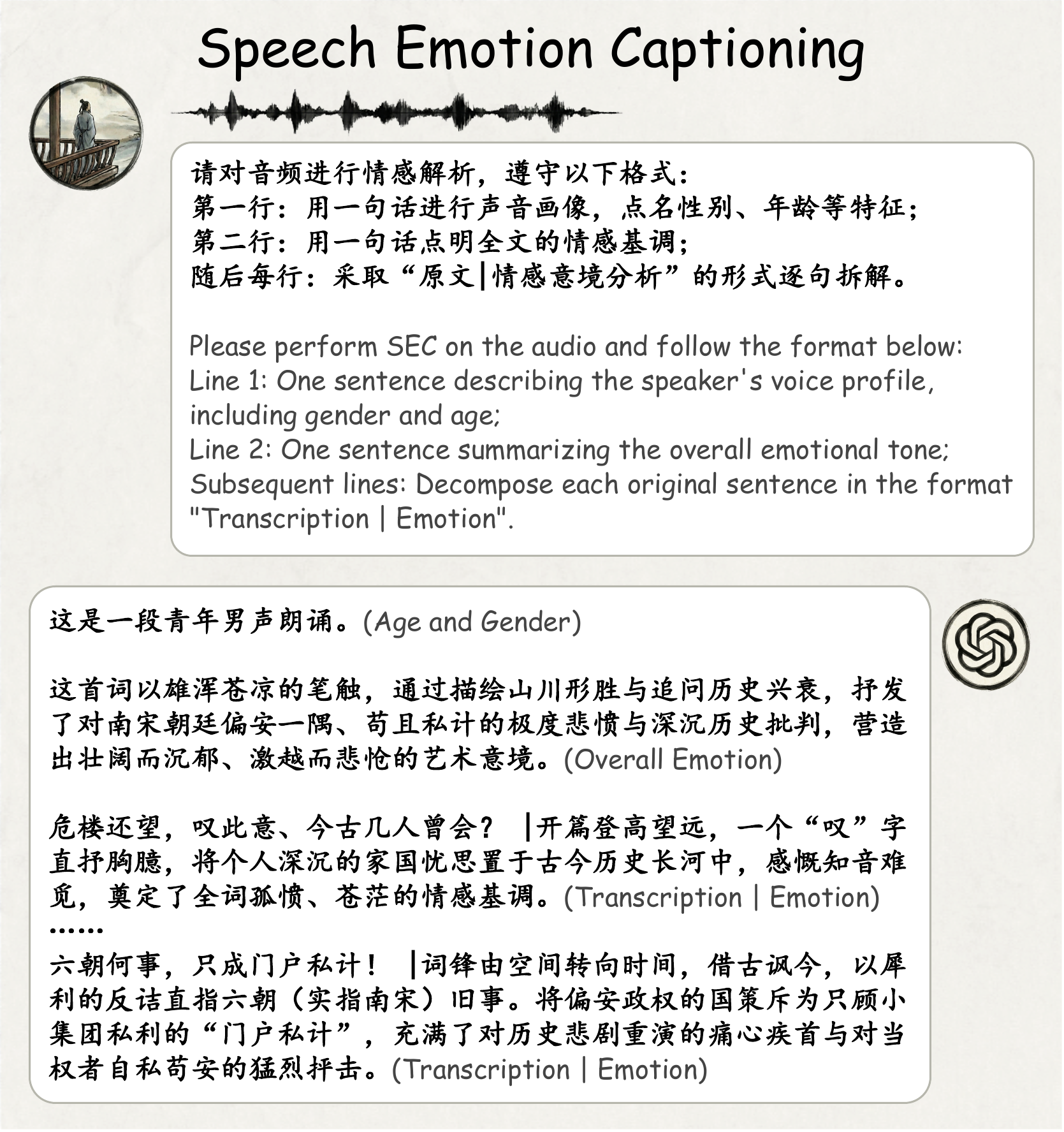}
\caption{\textbf{Case for SEC Task.} }
\label{case_sec}
\end{figure}

\begin{figure*}[t]
\centering
\includegraphics[width=\linewidth]{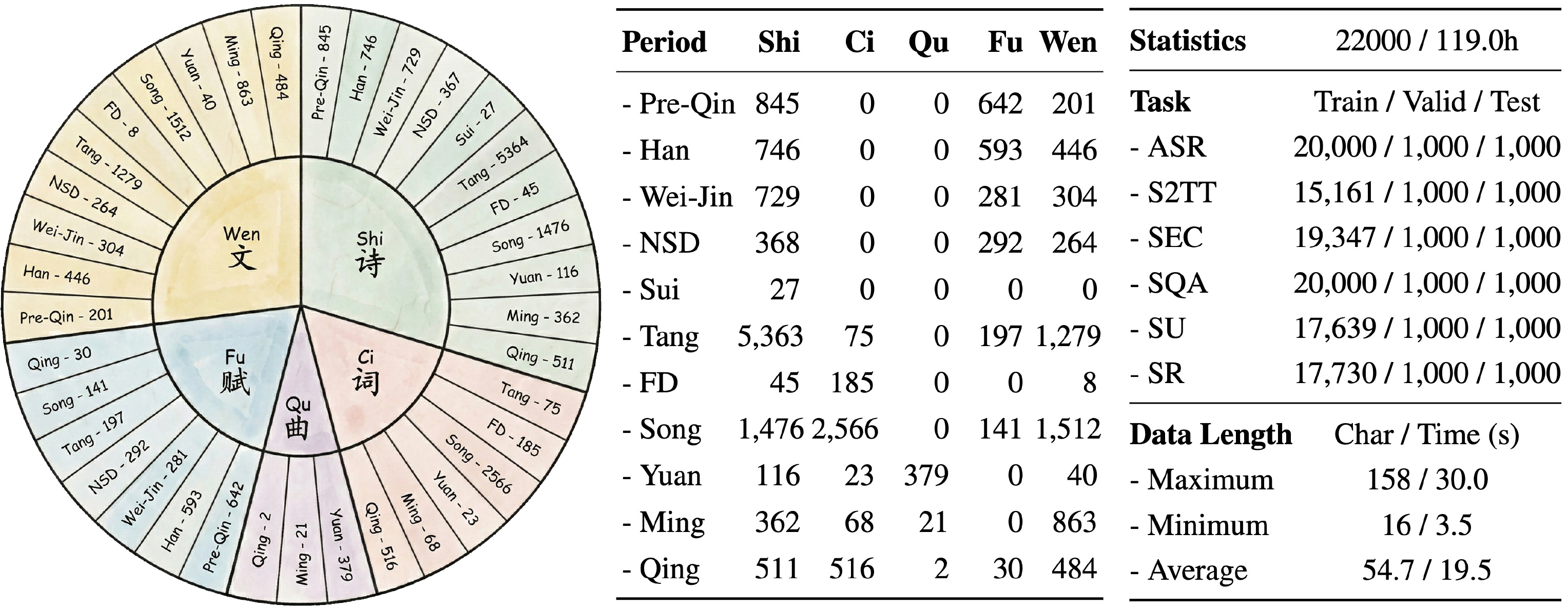}
\caption{\textbf{Corpus Statistics.}  It comprises 22,000 filtered human-recorded speech samples (totaling 119 hours) and supports 6 downstream tasks. Sample counts for S2TT, SEC, SU, and SR are lower due to the removal of invalid QA pairs. (NSD: the Northern and Southern Dynasties; FD: the Five Dynasties)}
\label{dataset}
\end{figure*}

\newpage

\subsection{Metric Definitions}
\paragraph{Emotion Caption Fidelity.} As shown in Figure~\ref{case_sec}, we design the Emotion Caption Fidelity (ECF) metric, an LLM-based score mechanism. The mechanism consists of the following parts:
\begin{itemize}
    \item \textbf{Persona Recognition (ECF-P, 0--2):} Measures the capability to extract identity features such as age and gender. Starting from an initial score of $2$, a penalty of $1$ point is deducted for each attribute error.
    \item \textbf{Global Emotional Tone (ECF-G, 0--3):} Evaluates the overall emotional atmosphere based on the richness and accuracy of the descriptions. A score of 0 is assigned if the emotional category or context is misidentified.
    \item \textbf{Sentence Emotion Fidelity (ECF-F, 0--5):} Evaluates sentence-by-sentence transcription and analysis. Deduct 1pt per emotional error and score 0 for hallucinations or irrelevant content.
\end{itemize}

\paragraph{Cross-modal Consistency.}
To evaluate how reliably MLLMs maintain consistency across different input modalities, we define the CMC metric as: 
\begin{equation}
\mathrm{CMC} = \frac{1}{3} \left( \frac{\mathrm{SQA}}{\mathrm{QA}} + \frac{\mathrm{SU}}{\mathrm{LU}} + \frac{\mathrm{SR}}{\mathrm{LR}} \right) \times 100
\label{eq:cmc}
\end{equation}
The CMC metric quantifies the performance consistency between speech and text inputs for the same set of questions by calculating the ratio of their respective scores.

\newpage

\subsection{Dataset Statistics}
Figure~\ref{dataset} shows the statistics of MCGA. It spans 5 genres across 11 historical periods, resulting in 37 unique period–genre categories.

Tang Shi has the most samples, followed by Song Ci. This is because the Tang and Song dynasties were the two peak periods of classical Chinese literature. Shi was the most popular genre in the Tang dynasty, while Ci was in the Song dynasty.

The corpus comprises 22,000 filtered human-recorded audio samples (totaling 119 hours) and supports 6 downstream tasks: ASR, S2TT, SEC, SQA, SU, and SR. The longest audio sample is 30 seconds, the shortest is 3.5 seconds, and the average duration is 19.5 seconds.

It should be noted that sample counts for S2TT, SEC, SU, and SR are lower due to the removal of invalid QA pairs. Also, the validation or test sets for the six tasks are not parallel.

\begin{table}[h]
  \centering
  \small
  \renewcommand{\arraystretch}{1.06} 
  \resizebox{\linewidth}{!}{
    \setlength{\tabcolsep}{8pt}
    \begin{tabular}{lcl} 
      \toprule 
      \textbf{Task} & \textbf{Metric} & \textbf{Details} \\ 
      \midrule
      \multirow{2}{*}{\textbf{ASR}}  & \multirow{2}{*}{CER $\downarrow$}      & {Text normalization} \\ 
      &&Following \citet{morris2004and}\\ \cmidrule(r){3-3}
      \multirow{2}{*}{\textbf{S2TT}} & \multirow{2}{*}{LLM-B $\uparrow$}     & LLM Evaluation \\ 
      &&{Following \citet{chen-etal-2025-benchmarking-llms}}\\ \cmidrule(r){3-3}
      \multirow{2}{*}{\textbf{SEC}}  & \multirow{2}{*}{ECF $\uparrow$}     & LLM Evaluation \\ 
       &&{Proposed in this work}\\  \cmidrule(r){3-3}
      \multirow{2}{*}{\textbf{SQA}}  & \multirow{2}{*}{F1 $\uparrow$}        & Open-ended \\ 
      &&{Factuality Evaluation}\\ \cmidrule(r){3-3} 
      \multirow{2}{*}{\textbf{SU}}   & \multirow{2}{*}{Accuracy $\uparrow$}  & Multiple-choice questions \\  
     && Options derived from the speech\\ \cmidrule(r){3-3}
      \multirow{2}{*}{\textbf{SR}}   & \multirow{2}{*}{Accuracy $\uparrow$}  & Multiple-choice questions \\  
     && External knowledge reasoning \\
      \bottomrule
    \end{tabular}
  }
  \caption{\textbf{Metric Details.} }
  \label{tab:metric}
\end{table}

\newpage

\section{Experiments}
\subsection{Experiment Setting}

\paragraph{Baseline MLLMs.} 
We evaluate 2 closed-source MLLMs ({GPT-4o-mini-Audio}~\cite{achiam2023gpt} and {Gemini-3-Flash}~\cite{gemini3flash}) and 8 open-source MLLMs: the {Qwen series}~\cite{chu2024qwen2,xu2025qwen2,xu2025qwen3}, the {Voxtral series}~\cite{liu2025voxtral}, {Phi-4-Multimodal-Instruct}~\cite{abouelenin2025phi}, {MiDashengLM}~\cite{dinkel2025midashenglm}, and {Step-Audio-2-mini}~\cite{wu2025stepaudio2technicalreport}.

\paragraph{Training Details.} We fine-tuned Qwen2.5-Omni-7B using the ms-swift framework\footnote{\url{https://github.com/modelscope/ms-swift}} with LoRA ($r=8, \alpha=32$)~\cite{hu2021loralowrankadaptationlarge}. The model was trained for 3 epochs on 4 A100 GPUs using the AdamW optimizer with a learning rate of $1 \times 10^{-4}$, a per-device batch size of 8, and a gradient accumulation of 4.

\paragraph{Evaluation Metrics.}
As shown in Table~\ref{genre}, we evaluate MLLMs across six tasks. All open-source models are deployed using the vLLM framework\footnote{\url{https://github.com/vllm-project/vllm}}~\cite{kwon2023efficient}, with inference performed via API requests at a temperature of 0. To provide a more intuitive performance metric, we normalize the S2TT and SEC results to a 100-point scale. Specifically, the ASR task is evaluated using the Character Error Rate (CER)\footnote{\url{https://github.com/jitsi/jiwer}}, while the S2TT and SEC tasks are scored by the deepseek-chat API~\cite{guo2025deepseek}. For SQA, we report the F1 score, and for SU and SR, we report Accuracy.

\begin{table*}[t]
\centering
\small
\renewcommand{\arraystretch}{1.3}
\resizebox{\textwidth}{!}{
\setlength{\tabcolsep}{6pt}
\begin{tabular}{l *{7}{w{c}{1cm}}}
\toprule
\multirow{2}{*}{\textbf{Model}} & \textbf{ASR} & \textbf{S2TT} & \textbf{SEC}& \textbf{SQA} & \textbf{SU} & \textbf{SR} & \multirow{2}{*}{\textbf{Score}~$\uparrow$}\\
& CER $\downarrow$ & LLM-B $\uparrow$ & ECF $\uparrow$ & F1 $\uparrow$ & Acc $\uparrow$ & Acc $\uparrow$ \\\midrule

 \multicolumn{7}{c}{\textbf{Closed-source MLLMs}} \\ \midrule

\rowcolor{my_green} GPT-4o-mini-Audio~\cite{achiam2023gpt} & 20.6 & 43.5 & 5.7 & 30.6 & 74.8 & 70.2 & 304.2\\
\rowcolor{my_green} Gemini-3-Flash~\cite{gemini3flash} &6.1 & \textbf{74.0} & 54.0 & 48.7 & 86.6& \textbf{83.7} &440.9\\ \midrule \multicolumn{7}{c}{\textbf{Open-source MLLMs}} \\ \midrule

\rowcolor{my_yellow} Phi-4-Multimodal-Instruct~\cite{abouelenin2025phi}   & 59.6 & 27.5 & 12.7 & 24.5 & 50.6 & 54.4 &210.1\\
\rowcolor{my_yellow} Voxtral-Mini~\cite{liu2025voxtral}      & 30.0 & 25.5 & 15.0 & 12.5 & 58.9 & 62.8 &244.7\\
\rowcolor{my_yellow} Voxtral-Small~\cite{liu2025voxtral}      & 30.0 & 34.1 & 16.4 & 27.9 & 72.6 & 71.9 &292.9\\

\rowcolor{my_yellow} MiDashengLM~\cite{dinkel2025midashenglm}       & 11.7 & 42.9 & 24.7 & 22.5 & 72.2 & 75.6 &326.2\\
\rowcolor{my_yellow} Step-Audio-2-mini~\cite{wu2025stepaudio2technicalreport}   & 9.9& 41.9&36.8&45.2&80.5&80.4&374.9\\
\rowcolor{my_yellow} Qwen2-Audio-7B-Instruct~\cite{chu2024qwen2}   & 18.7&30.4&26.1&24.8&72.1&64.7&299.4\\
\rowcolor{my_yellow} Qwen2.5-Omni-7B~\cite{xu2025qwen2}   & 10.1 & 49.7 & 37.0 & 43.5 & 81.3 & 79.3&380.7\\
\rowcolor{my_yellow} Qwen3-Omni-30B-A3B-Instruct~\cite{xu2025qwen3}     & \textbf{4.4}  & 67.6 & \textbf{58.4} & \textbf{51.5} & \textbf{86.9} &82.9 &\textbf{442.9}\\
\bottomrule
\end{tabular}}
\caption{\textbf{Performance Comparison of Different MLLMs on the MCGA Test Set.} Detailed results for LLM-B and ECF are shown in Table~\ref{tab:s2tt} and Table~\ref{sec}. The overall Score is the aggregate of the six tasks, where the ASR component is calculated as $100 - \text{CER}$.}
\label{tab:main}
\end{table*}

\newpage

\subsection{Main Results}
We present a comprehensive evaluation of ten MLLMs across six audio tasks. By analyzing the interplay between model performance and task difficulty, we derive the following key observations:

\paragraph{Closed-source vs. Open-source Models.} 
In Table~\ref{tab:main}, Qwen3-Omni demonstrates superior performance on the MCGA test set for Chinese understanding and generation tasks, specifically in \textbf{ASR} (4.4 CER $\downarrow$), \textbf{SEC} (58.4 LLM-C $\uparrow$), \textbf{SQA} (51.5 F1 $\uparrow$), and \textbf{SU} (86.9 Acc $\uparrow$). Conversely, closed-source models such as Gemini-3-Flash maintain a competitive edge in English generation and Chinese reasoning tasks, leading in metrics such as \textbf{S2TT} (74.0 LLM-B $\uparrow$) and \textbf{SR} (83.7 Acc $\uparrow$). Overall, the open-source models have achieved a competitive level of performance compared with the closed-source ones.

\begin{figure*}[h]
\centering
\includegraphics[width=\linewidth]{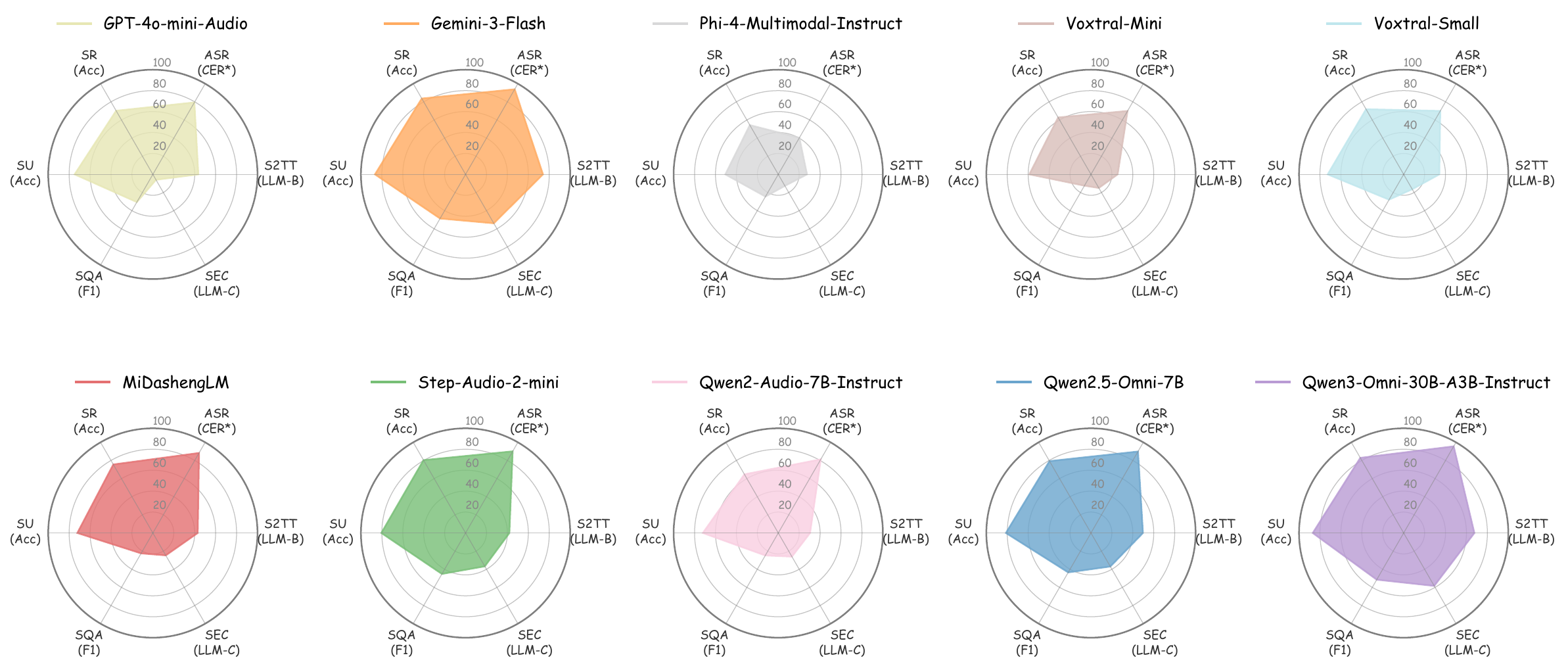}
\caption{\textbf{Comparison across Different Tasks.} Existing MLLMs exhibit robust performance in \textbf{ASR}, \textbf{SU}, and \textbf{SR} tasks, but they still encounter challenges regarding the beauty of translation in \textbf{S2TT}, affective modeling in \textbf{SEC}, and hallucination issues in open-ended \textbf{SQA}. CER$^*$ refers to $(100 - \text{CER})$.}
\label{task}
\end{figure*}

\paragraph{Comparison across Different Tasks.}
As shown in Figure~\ref{task}, existing MLLMs demonstrate their strongest performance in CCS \textbf{ASR} tasks. This is followed by \textbf{SU} and \textbf{SR} in multiple-choice formats, where models achieve relatively robust results. Regarding the \textbf{S2TT} task, overall performance is acceptable but remains to be enhanced. In contrast, performance on \textbf{SEC} is notably poor, indicating a critical need for enhanced affective computing capabilities. Finally, for \textbf{SQA}, F1 scores remain low, suggesting that "hallucination" issues~\cite{du2025ccfqabenchmarkcrosslingualcrossmodal} have yet to be effectively resolved.

\newpage

\subsection{Further Analysis}

\subsubsection{Analysis of ASR Task}

\paragraph{Performance Disparity Across Genres.}
Table~\ref{genre} shows the MLLMs' performance which varies significantly by genre. Qwen3-Omni achieves state-of-the-art results on MCGA, maintaining the lowest CER across all categories, particularly in \textit{Ci} (2.8). A consistent trend across various models is that \textit{Ci} achieves lower CER, while \textit{Fu} consistently poses the greatest difficulty. This difficulty stems from \textit{Fu}'s ornate rhetoric, frequent classical allusions, and high density of modal particles.
\begingroup
\begin{table}[h]
\centering
\small
\renewcommand{\arraystretch}{1.2} 
        \setlength{\tabcolsep}{2pt} 
\resizebox{\linewidth}{!}{
\begin{tabular}{lccccc}
\toprule
\multirow{-1}{*}{\textbf{Models}}                                              &  \multirow{-1}{*}{\textbf{Shi}}          &   \multirow{-1}{*}{\textbf{Ci}}              &    \multirow{-1}{*}{\textbf{Qu}}              &    \multirow{-1}{*}{\textbf{Fu}}                                       &       \multirow{-1}{*}{\textbf{Wen}}    \\
 \midrule
\rowcolor{my_green} GPT-4o-mini-Audio   &22.6&20.3&19.0&22.8&\underline{18.5} \\
\rowcolor{my_green} Gemini-3-Flash  & \underline{6.1}&6.8&7.7&8.4&\underline{6.1}\\ \midrule
\rowcolor{my_yellow}  Phi-4-Multimodal-Instruct &61.0&63.8&63.0&61.1&\underline{51.1}\\
\rowcolor{my_yellow} Voxtral-Mini &29.2&\underline{27.4}&29.2&34.5&28.9\\
\rowcolor{my_yellow} Voxtral-Small  &30.6&\underline{26.4}&30.1&33.9&28.4\\
\rowcolor{my_yellow} MiDashengLM  &12.7&10.1&\underline{9.4}&15.7&10.0 \\
\rowcolor{my_yellow} Step-Audio-2-mini &9.0&\underline{7.0}&7.5&14.6&10.3\\
\rowcolor{my_yellow} Qwen2-Audio-7B-Instruct &18.2&\underline{15.8}&{15.9}&23.3&19.0\\ 
\rowcolor{my_yellow} Qwen3-Omni-30B-A3B-Instruct&\textbf{3.8}&\textbf{\underline{2.8}}&\textbf{4.1}&\textbf{6.2}&\textbf{4.3}\\ \midrule
\rowcolor{my_yellow} Qwen2.5-Omni-7B &9.9&\underline{7.5}&8.9&14.8&8.8\\ 
\rowcolor{my_yellow} Qwen-Omni-MCGA &\underline{\textbf{2.8}}&{3.1}&7.8&\textbf{5.3}&\textbf{4.1}\\ 
\bottomrule
\end{tabular}}
\caption{\textbf{CER Scores Across Different Genres.} The test set contains 1,000 samples (200 per genre). \underline{Underline} indicates the best-performing genre for each individual model. Qwen-Omni-MCGA is a LoRA-based adaptation of Qwen2.5-Omni-7B. It achieves state-of-the-art results on all genres except for \textit{Qu}.}
\label{genre}
\end{table}
\endgroup

\begin{figure}[h]
\centering
\includegraphics[width=1\linewidth]{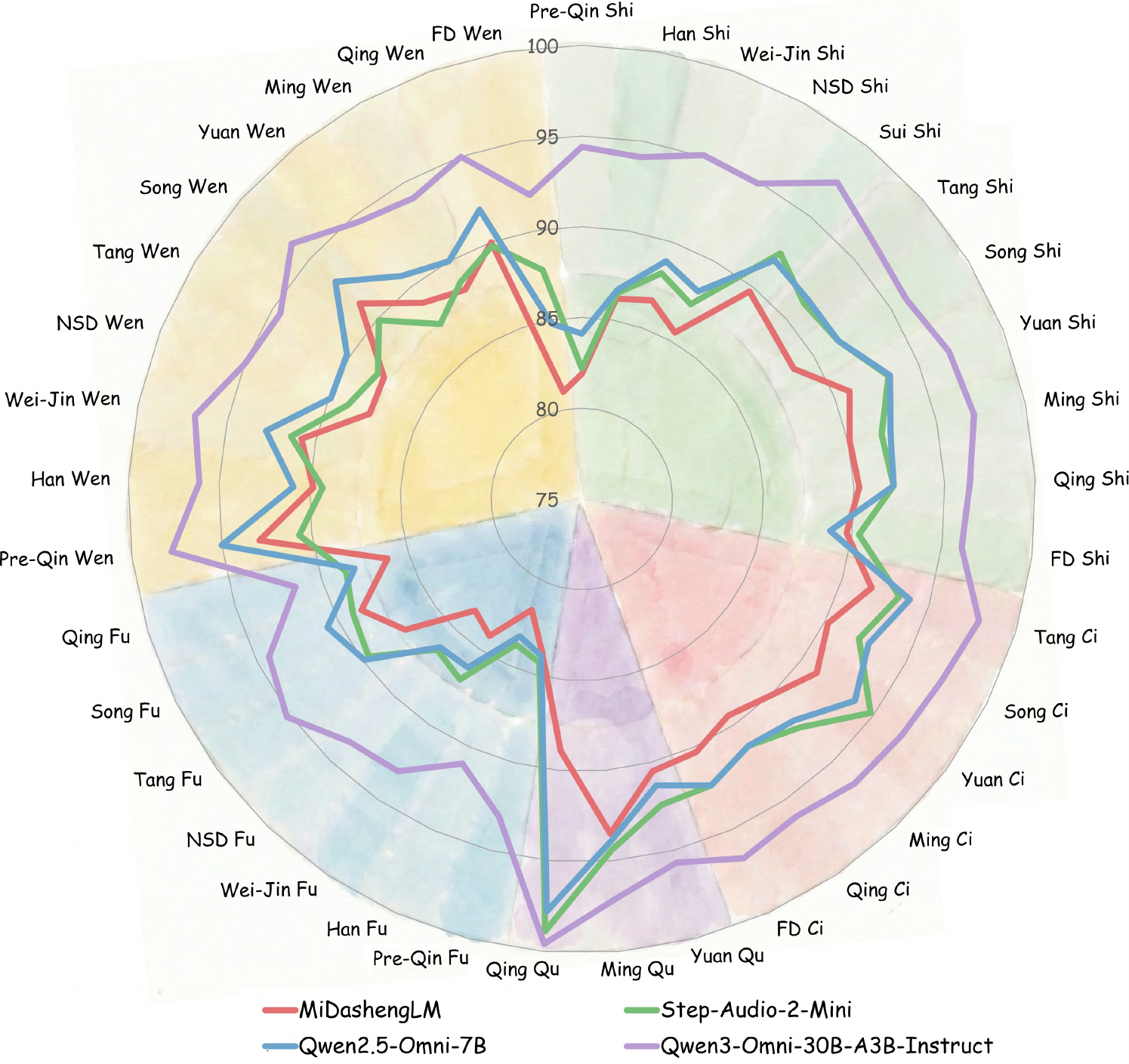}
\caption{\textbf{CER$^*$ Across Dynasties and Genres.}}
\label{dynasty}
\end{figure}

\paragraph{Audio Quality.} Table~\ref{cer_qa} reveals a 0.1 CER gap (Qwen3-Omni) between human-verified valid/test sets and the train set, confirming high data consistency. Residual errors primarily stem from uncommon characters and phonetic loanwords (\textit{tongjiazi}) in Classical Chinese. Figure~\ref{dynasty} shows the CER distribution across dynasties and genres.

\begingroup
\renewcommand{\arraystretch}{1.0} 
        \setlength{\tabcolsep}{6pt} 
\begin{table}[h]
\centering
\small
\resizebox{\linewidth}{!}{
\begin{tabular}{lcccc}
\toprule

\multirow{-1}{*}{\textbf{Models}}                                             &       \multirow{-1}{*}{\textbf{Train}}          &   \multirow{-1}{*}{\textbf{Valid}}               &    \multirow{-1}{*}{\textbf{Test}}                                       &       \multirow{-1}{*}{\textbf{Avg.}}           \\

\midrule
\rowcolor{my_yellow} MiDashengLM  &11.7&11.2&11.7&11.7\\
\rowcolor{my_yellow}Step-Audio-2-mini &10.4&10.1&9.9&10.4\\
\rowcolor{my_yellow}Qwen2.5-Omni-7B &9.8&9.5&10.1&9.8\\
\rowcolor{my_yellow}Qwen3-Omni-30B-A3B-Instruct &4.5&4.4&4.4&4.5\\

\bottomrule
\end{tabular}}
\caption{\textbf{CER Scores for Quality Check.} The train, valid, and test sets show high data consistency.}
\label{cer_qa}
\end{table}
\endgroup

%


\subsubsection{Analysis of S2TT Task}

\paragraph{Beauty Evaluation of Translation.} 
As illustrated in Table 6, we evaluate the translation quality across four dimensions: Beauty of Form (LLM-BF), Beauty of Meaning (LLM-BM), Beauty of Sound (LLM-BS), and their average score (LLM-B). The closed-source model \textbf{Gemini-3-Flash} achieves the highest performance across all metrics, reaching a peak average score of 74.0 (LLM-B $\uparrow$). 

\paragraph{S2TT Quality.} 
Additionally, we provide high-quality ground-truth translation candidates. The LLM-B score of 79.2 (4.0) is constrained by the 1--5 evaluation scale, as the DeepSeek API evaluation model typically assigns moderate scores and rarely grants a perfect score of 5. To provide a more intuitive performance metric, we normalize these raw API scores to a 100-point scale.

\begin{table}[h]
\centering
\small
\renewcommand{\arraystretch}{1.2}
\setlength{\tabcolsep}{4pt}
\resizebox{\linewidth}{!}{
\begin{tabular}{lccccc}
\toprule
\textbf{Models} & \textbf{COMET} & \textbf{BF} & \textbf{BM} & \textbf{BS} & \textbf{LLM-B} \\ \midrule

\rowcolor{my_green}GPT-4o-mini-Audio & 53.3 & 42.3 & 42.8 & 45.3 & 43.5 \\
\rowcolor{my_green}Gemini-3-Flash & \textbf{58.9} & \textbf{72.4} & \textbf{74.5} & \textbf{75.2} & \textbf{74.0} \\
\midrule
\rowcolor{my_yellow}Phi-4-Multimodal-Instruct & 34.2 & 27.0 & 27.3 & 28.1 & 27.5 \\
\rowcolor{my_yellow}Voxtral-Mini & 46.9 & 25.6 & 25.6 & 25.1 & 25.5 \\
\rowcolor{my_yellow}Voxtral-Small & 52.0 & 34.5 & 34.2 & 33.6 & 34.1 \\
\rowcolor{my_yellow}MiDashengLM & 47.0 & 43.8 & 44.3 & 40.6 & 42.9 \\
\rowcolor{my_yellow}Step-Audio-2-mini & 53.7 & 42.6 & 43.3 & 39.6 & 41.9 \\
\rowcolor{my_yellow}Qwen2-Audio-7B-Instruct & 38.2 & 29.8 & 30.6 & 30.9 & 30.4 \\
\rowcolor{my_yellow}Qwen2.5-Omni-7B & 55.1 & 51.3 & 51.7 & 46.1 & 49.7 \\
\rowcolor{my_yellow}Qwen3-Omni-30B-A3B-Instruct & 58.7 & 68.1 & 69.1 & 65.5 & 67.6 \\
\bottomrule
\end{tabular}
}
\caption{\textbf{Beauty Evaluation of Translation.} Following \citet{chen-etal-2025-benchmarking-llms}, we employ COMET~\citep{rei2020comet} and Beauty metrics: Beauty of Form (BF), Beauty of Meaning (BM), and Beauty of Sound (BS). LLM-B denotes the mean of the three beauty metrics.}
\label{tab:s2tt}
\end{table}

\newpage

\subsubsection{Analysis of SEC Task}

\paragraph{Open-source vs. Closed-source MLLMs.} 
As shown in Table~\ref{sec}, Qwen3-Omni outperforms other models across all ECF metrics. This superior performance is attributed to its deep understanding of Chinese cultural nuances and its robust transcription capabilities. It is followed by Gemini-3-Flash, which maintains competitive results.

In contrast, GPT-4o-mini-Audio exhibits poor performance. This is primarily because its stringent safety protocols frequently trigger refusals when tasked with persona-based or emotional analysis.

\begingroup
\begin{table}[h]
\centering
\small
\renewcommand{\arraystretch}{1.2} 
        \setlength{\tabcolsep}{2pt} 
\resizebox{\linewidth}{!}{
\begin{tabular}{lcccc}
\toprule
\multirow{-1}{*}{\textbf{Models}}                                              &   \multirow{-1}{*}{\textbf{ECF-P}}              &    \multirow{-1}{*}{\textbf{ECF-G}}              &    \multirow{-1}{*}{\textbf{ECF-S}}                                       &       \multirow{-1}{*}{\textbf{ECF}}           \\ \midrule
Ground Truth&20.0&30.0&50.0&100.0\\
 \midrule
\rowcolor{my_green}GPT-4o-mini-Audio  & 1.4&3.2&1.2&5.7\\
\rowcolor{my_green}Gemini-3-Flash  & 13.4&16.9&23.6&54.0\\ \midrule
\rowcolor{my_yellow}Phi-4-Multimodal-Instruct &4.3&7.9&0.5&12.7\\
\rowcolor{my_yellow}Voxtral-Mini &5.5&8.9&0.6&15.0\\
\rowcolor{my_yellow}Voxtral-Small  &2.2&11.9&2.3&16.4\\
\rowcolor{my_yellow}MiDashengLM  &13.6&8.0&3.2&24.7 \\
\rowcolor{my_yellow}Step-Audio-2-mini &16.2&11.8&8.8&36.8\\
\rowcolor{my_yellow}Qwen2-Audio-7B-Instruct &12.9&10.4&2.7&26.1\\
\rowcolor{my_yellow}Qwen2.5-Omni-7B &14.3&13.9&8.8&37.0\\ 
\rowcolor{my_yellow}Qwen3-Omni-30B-A3B-Instruct&\textbf{16.0}&\textbf{18.6}&\textbf{23.8}&\textbf{58.4}\\ 
\bottomrule
\end{tabular}}

\caption{\textbf{LLM-based Evaluation for SEC.} 
(1) ECF-P ($0\text{--}2$) for persona identification; 
(2) ECF-G ($0\text{--}3$) for global emotional tone analysis; 
(3) ECF-S ($0\text{--}5$) for sentence-level emotion; 
(4) ECF is the sum of scores.}
\label{sec}
\end{table}
\endgroup

\subsubsection{Analysis of SQA, SU, and SR Tasks}

\paragraph{Open-ended vs. Multiple-choice QA.} 
As shown in Table \ref{cmc}, a substantial performance gap exists between multiple-choice and open-ended formats. MLLMs struggle significantly more with open-ended questions, such as identifying authors or titles, compared to complex reasoning tasks that provide candidate options. For instance, Gemini-3-Flash scored 86.6 in SU and 83.7 in SR but drops to 48.7 in SQA. This gap indicates that MLLMs suffer from severe hallucinations in open-ended factual QA, despite their strong reasoning.

\paragraph{Cross-modal Consistency.}

As shown in Table \ref{cmc}, SQA, SU, and SR represent performance on speech-based tasks, while the denominators QA, LU (Language Understanding), and LR (Language Reasoning) serve as the text-only upper-bound references. Step-Audio-2-mini achieved the highest CMC score among all evaluated MLLMs.

\begingroup
\begin{table}[h]
\centering
\small
\renewcommand{\arraystretch}{1.2} 
        \setlength{\tabcolsep}{1pt} 
\resizebox{\linewidth}{!}{
\begin{tabular}{lccccccc}
\toprule
\multirow{-1}{*}{\textbf{Models}}                                              &   \multirow{-1}{*}{\textbf{SQA}}              &    \multirow{-1}{*}{\textbf{SU}}              &    \multirow{-1}{*}{\textbf{SR}}                                       &       \multirow{-1}{*}{\textbf{QA}}    &       \multirow{-1}{*}{\textbf{LU}}  &       \multirow{-1}{*}{\textbf{LR}}   & \multirow{-1}{*}{\textbf{CMC}}   \\
\midrule
\rowcolor{my_green}Gemini-3-Flash  & {48.7}&{86.6}&\textbf{83.7}&\textbf{ 66.0}&\textbf{94.6}&\textbf{91.5}&{85.6}\\ \midrule
 \rowcolor{my_yellow}Phi-4-Multimodal-Instruct &24.5&50.6&54.4&25.2&69.1&60.4&86.8\\
\rowcolor{my_yellow}Voxtral-Mini &12.5&58.9&62.8&13.7&76.8&69.5&86.1\\
\rowcolor{my_yellow}Voxtral-Small  &27.9&72.6&71.9&40.0&88.8&83.1&79.3\\
\rowcolor{my_yellow}MiDashengLM  &22.5&72.2&75.6&45.1&89.1&84.4 &73.5\\
\rowcolor{my_yellow}Step-Audio-2-mini &45.2&80.5&80.4&51.3&89.6&85.4&\textbf{90.7}\\
\rowcolor{my_yellow}Qwen2-Audio-7B-Instruct &24.8&72.1&64.7&36.0&79.9&70.7&83.5\\ 
\rowcolor{my_yellow}Qwen2.5-Omni-7B &43.5&81.3&79.3&53.1&91.6&85.0&88.0\\ 
\rowcolor{my_yellow}Qwen3-Omni-30B-A3B-Instruct&\textbf{51.5}&\textbf{86.9}&82.9&60.0&93.6&91.0&89.9\\  
\bottomrule
\end{tabular}}
\caption{\textbf{Cross-modal Consistency.} CMC quantifies the performance gap between audio and textual modalities.}
\label{cmc}
\end{table}
\endgroup

\section{Conclusion}

This paper presents MCGA, the first large-scale, fully copyrighted audio corpus dedicated to CCS, encompassing six tasks. Besides, we propose the emotion caption fidelity metric for SEC task alongside a metric for cross-modal consistency. Our systematic evaluation of ten MLLMs reveals that the Qwen series leads in performance, exhibiting superior proficiency in comprehending classical Chinese literature.

\section{Limitations}
Although MCGA incorporates audio-text multimodal data across six distinct tasks, several limitations persist. First, copyright constraints preclude the inclusion of real-world image samples that are precisely aligned with both textual and auditory modalities. Second, the \textit{Qu} genre emerged significantly later than \textit{Shi}, \textit{Ci}, \textit{Wen}, and \textit{Fu}. Due to the relatively short-lived nature of the Yuan Dynasty, the volume of extant works is considerably limited, leading to a lower representation of \textit{Qu} within the corpus.

\section{Ethical Considerations}
Ethical integrity is central to our research involving human audio data. All data utilized in this study were sourced from volunteers who contacted the authors directly. Volunteers received equitable compensation and signed a Voice Authorization License Agreement, granting unambiguous permission for their speech to be used in a research context. We ensure that data handling strictly aligns with applicable privacy laws and data protection mandates. To protect participant privacy, all entries in the final corpus have been anonymized.

\section*{Acknowledgements}
The research in this article is supported by the National Science and Technology Major Program (Grant No. 2024ZD01NL00101), the National Science Foundation of China (U22B2059, 62276083, 62506182), National Key Research and Development Program of China (2025YFE0200500), the Key Research and Development Program of Heilongjiang Province (2022ZX01A28) and the 5G Application Innovation Joint Research Institute’s Project (A003), and the Major Key Project of PCL (Grant No. PCL2025A12, PCL2025A03).

\bibliography{custom}

\end{document}